\documentclass[conference]{IEEEtran}
\IEEEoverridecommandlockouts

\usepackage{cite}
\usepackage{amsmath,amssymb,amsfonts,setspace}
\usepackage{algorithmic}
\usepackage{graphicx}
\usepackage{textcomp}
\usepackage{xcolor}

\usepackage{tikz, tikz-cd, graphics, graphicx, mathtools, amsfonts, mathrsfs, amsmath, xfrac, dsfont, enumitem, multirow, mathdots}

\usepackage[ruled,vlined]{algorithm2e}
\SetKwInOut{Parameters}{Parameters}
\SetKwInOut{Initialization}{Initialization}
\usepackage{orcidlink}
\usepackage{stfloats}

\usepackage{cite}
\usepackage{amsmath,amssymb,amsfonts}
\usepackage{algorithmic}
\usepackage{graphicx}
\usepackage{textcomp}
\usepackage{xcolor}
\def\BibTeX{{\rm B\kern-.05em{\sc i\kern-.025em b}\kern-.08em
    T\kern-.1667em\lower.7ex\hbox{E}\kern-.125emX}}
\begin{document}

\title{A Multi-Head Attention Approach for SLA Compliance Monitoring in Data Centers\\
\thanks{Accepted for Publication in the 46th IEEE International Conference on Distributed Computing Systems.
Aspects of this work are the subject of a pending patent application.}
}

\author{\IEEEauthorblockN{Omanshu Thapliyal \orcidlink{0000-0003-3847-188X}}
\IEEEauthorblockA{\textit{Researcher, Strategic Data Solutions Lab} \\
\textit{Hitachi America Ltd.}\\
Santa Clara, USA }
}
\maketitle

\begin{abstract}
Service level agreements (SLAs) in data center colocation contracts define precise thresholds for power, temperature, and humidity, with tiered violation penalties expressed as credits against monthly recurring charges. 
Traditional reactive monitoring detects breaches only after they occur, limiting remediation opportunities. 
We present a framework that encodes SLA rules as structured JSON objects
to generate training data without manual annotation. 
We train a per-customer multi-head transformer model in which each attention head specializes in one SLA rule, learning temporal dependencies that precede violations by 30 minutes. 
Post-training, the inference service emits structured prediction events transformed into three role-specific views: finance schemas exposing credit liability, operations schemas surfacing risk scores and recommended interventions, and compliance schemas bundling predictions with immutable telemetry signatures for audit. 
By aligning model architecture directly with contractual obligations, this framework enables operators to anticipate SLA breaches, prioritize corrective actions, and minimize financial penalties.
\end{abstract}


\section{Introduction}
Colocation data centers (or colos) are large-scale facilities that rent out space to host their servers, storage, and networking hardware to cater to enterprise customers.
Unlike traditional hyperscaler data centers that are owned, managed, and occupied by bigger tech enterprises, colos are operated by third-party providers and designed for multi-tenant, shared operations. 
As the industry shifts towards increasing artificial intelligence (AI) workloads, the tenant requirements for resilient, compliant, persistently available, and scalable infrastructure have increased significantly.
While training large models takes place in the massive hyperscaler facilities, inference tasks are pushing enterprise tenants to adopt colo models at increasing rates. 
Since inference requires low latency and tenant proximity, businesses are leasing colo spaces to run ``edge" AI models closer to their data sources and customers.

As a result of this industry shift, reports project a 160-175\% increase in global data center power demand by 2030 \cite{sachs2025ai}, with AI workloads comprising roughly half of all digital infrastructure demands \& compute workloads \cite{mills2025rise}.
The colocation industry is expected to experience a resulting CAGR of over 35\%, compared to the broarder AI data center industry CAGR of 31\% \cite{marketsandmarkets2025aidatacenter, shehabi20242024, davenport2024ai}.
Additional industry drivers in the form of the increased 1-year cadence for new GPU release cycles from NVIDIA, a push towards $\pm$400V direct current, direct-to-chip liquid cooling, and 100+kW per rack infrastructure, the total global power demand for the colo sector is expected to exceed 150-200 GW by 2030 \cite{sachs2025ai, shehabi20242024}.

From a colo perspective, the increasing push towards AI-intensive workloads means that tenants not only expect space, power, and cooling, but also provable adherence to stringent Service Level Agreements (SLAs).
These SLAs cover uptime, power availability, environment stability, and support responsiveness, while outlining remedies like service credits when these contracted agreements may not be met.
These SLAs also exclude downtime caused by scheduled maintenance, client-side equipment failure, or \textit{force majeure} events.

For a colocation provider, the SLAs are both a competitive differentiator and a financial risk instruments.
Operators typically guarantee availability in the 99.9-99.999\% ranges (often called "3-9's and 5-9's in the industry"), translating to mere minutes of downtime per month.
These guarantees are backed by service credits translating directly to dollar amounts, upon the breach of agreed thresholds.
Conversely, enterprise customers evaluate colocation partners based on SLA transparency, auditability, monitoring capability, and past demonstrations of SLA compliance. 
One such compliance failure took place in 2025, when a 10-hour global trading halt was triggered for CME Group in Illinois, due to a mechanical failure in the cooling systems at the Chicago area facility \cite{reuters2025cyrusone}. 
This outage amounted to \$1.58 billion in lost trading volume for the CME Group, highlighting the need for real-time SLA monitoring. 

Effective SLA compliance monitoring for colocation facilities therefore must satisfy several competing goals:
\begin{itemize}
\item Making contractual agreements machine-readable: SLA documents are often unstructured, natural language contracts and agreements, with tenant and facility specific rules that may be executed consistently across facilities.
\item Shifting from reactive to proactive risk management: As-is colo operations involve reactive management of SLA rules upon the occurrence of a breach event. This involves flagging violations \textit{after they happen}. The system should predict likelihoods and severity of breaches, giving operators time to intervene in breach events \textit{before they happen}.
\item Providing auditable, stakeholder-specific views: Finance, government, healthcare, legal customers have different needs for tailored outputs, as do finance, operations, legal team for colo operations. Credits issued, logs, key performance indicators (KPIs), and violation history should provide a trace to the same underlying rules, sensor data from physical systems, and the associated breach events. 
\end{itemize}

To this end, in this paper, we describe an SLA compliance monitoring architecture to address the above needs for colocation providers by combining the following.
We propose a ReAct-based agentic framework that ingests and interprets tenant SLAs into a structured rules database (DB). 
The associated structured rules DB is formed after anonymizing SLA documents, stripping them of all customer sensitive information.
Next, a per-customer, multi-head transformer model is trained to predict SLA breach events from past power, temperature, and humidity telemetry data. 
The result is an end-to-end contract-aware SLA monitoring pipeline designed for multi-tenant colocation facilities. 
Each customer's SLA is modeled explicitly, and rules are enforced in real-time conditions. 
Upcoming breach events are predicted for risk windows with enough lead time to avoid penalties and protect financial exposure from operational activities.

The remainder of this paper is organized as follows.
In Section \ref{sec:solution}, we provide a detailed outline of the solution framework.
In Section \ref{sec:attention}, we detail the multi-headed attention model for SLA rules monitoring specific to colocation operations.
Section \ref{sec:demo} contains our initial experimental results and a demonstration of the end-to-end system.
Finally, in Section \ref{sec:conclusion} we present our concluding remarks and future works.
The end-to-end framework control flow ensures a closed loop where contractual obligations defined on paper directly drive the interpretation of sensor/telemetry data, operational \& financial risk calculations, and allows direct modifications to SLA contracts based on structured SLA rules database (DB) after the occurrences of (or predicted) breach events.

\section{End-to-End SLA Monitoring Framework}\label{sec:solution}

\begin{figure*}[b]
    \centering
    \includegraphics[width=0.775\textwidth]{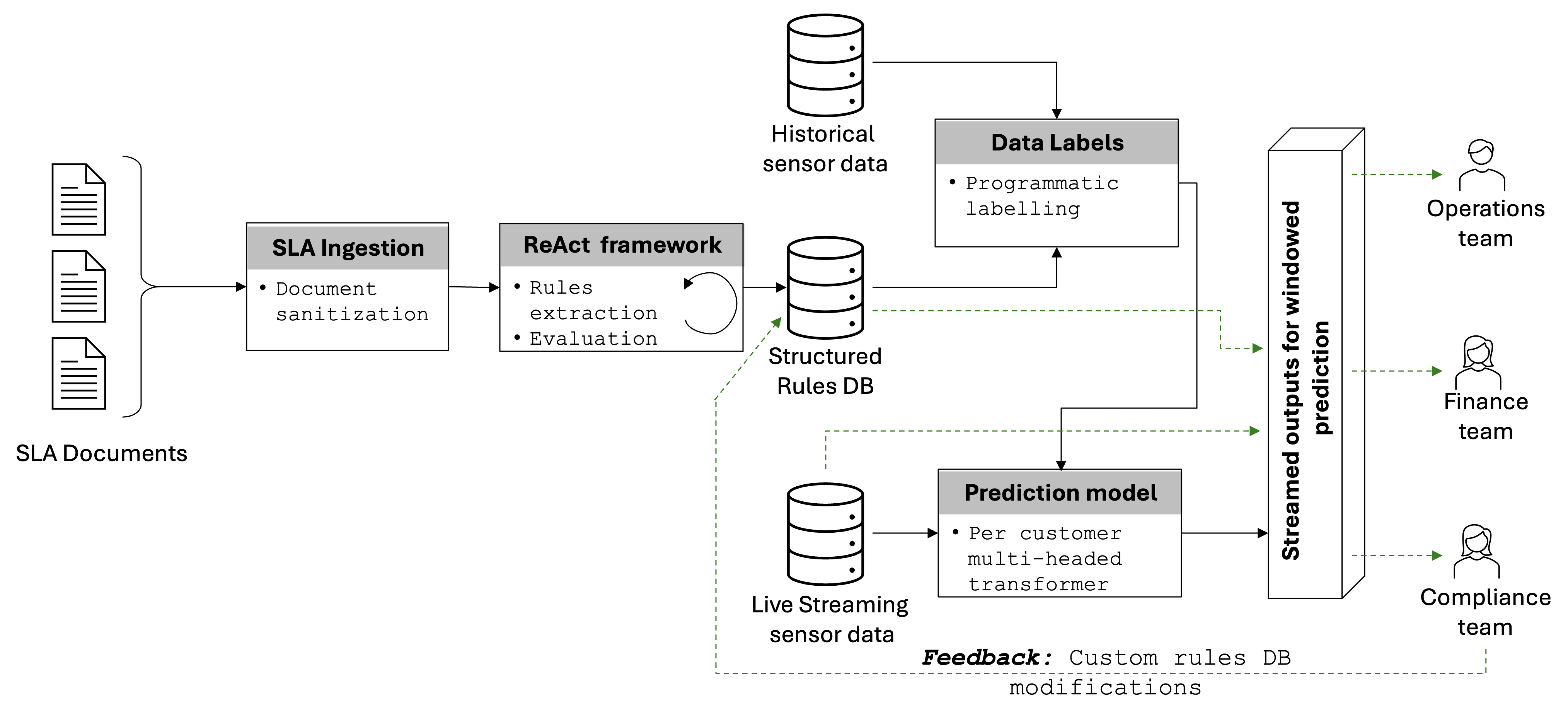}
    \caption{Architecture for Real-Time SLA Compliance Monitoring in High-Density Colocation Data Centers.}
    \label{fig:framework}
\end{figure*}

This section describes the high level solution architecture for our proposed End-to-end SLA Monitoring Framework.
On a high level, the framework consists of four main systems, shown in Fig.~\ref{fig:framework} as: SLA Ingestion system, rules extraction ReAct framework system, a programmatic data labeling system, and finally the per-customer multi-headed attention model for SLA violation prediction.

The framework performs 3 high level functions: 
\begin{itemize}
\item turns tenant SLA documents into a machine-readable, structured rules DB
\item uses the rules DB to programmatically label and interpret colocation data center telemetry data, and 
\item uses a per-customer multi-head attention model to predict SLA violation events and feeds operational, financial, and compliance outputs to relevant stakeholders.
\end{itemize}
This is scoped for a colocation setting where multiple customers share infrastructure but keep logically isolated SLAs and monitoring, often outfitting the facilities and racks with custom sensors for their own telemetry records.

\begin{figure}[h!]
    \centering
\includegraphics[width=1\columnwidth]{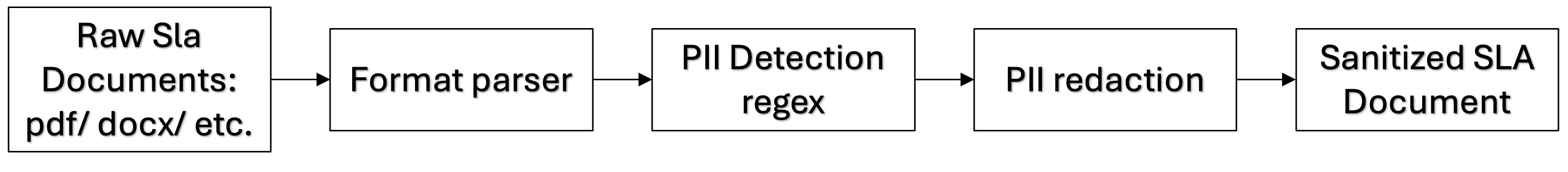}
    \caption{Ingestion \& PII Stripping System}
    \label{fig:system1}
\end{figure}

The ingestion \& privacy system in Fig.~\ref{fig:framework} parses heterogeneous contract files, that are contracts usually maintained in natural language, in PDF, DOCX, other formats.
As shown in Fig.~\ref{fig:system1}, at the entry point, the system parses common enterprise document formats and passes them through a format parser that standardizes text encoding and preserves sections, clause boundaries, and most importantly, recovers SLA tables and/or bullet point structures wherever they exist across the documents.
This normalization is critical for later ReAct agents downstream, which rely on consistent structure in the parsed data (headings, bullet points, tabular rows, and definitions).
Next, customer sensitive and personally identifiable information (PII), such as names, signatures, and direct contact information, tenant specific identifiers, etc. are scrubbed. 
This is done via \texttt{re} modules in Python, using regular expressions (Regex). 
Detected entities are either removed, or replaced with stable placeholders to preserve contractual semantics after scrubbing (e.g., \texttt{CME Corp, Chicago-1} is replaced with \texttt{Customer\_A, Facility\_1}).
This is in line with emerging practices in PII redaction for prompts that later are utilized in large language model (LLM) pipelines \cite{melnyk56safe, nakka2024pii}, where an isolated redaction later ensures sensitive data never leaves the controlled infrastructure while preserving correct semantics downstream. 
The output at this state is a "sanitized SLA document" objects, which is a structured artifact that bundles normalized text, sections, table clauses, and derived metadata, acting as the sole interface exposed to the subsequent system.

\begin{figure}[h!]
\centering
    \includegraphics[width=0.6\columnwidth]{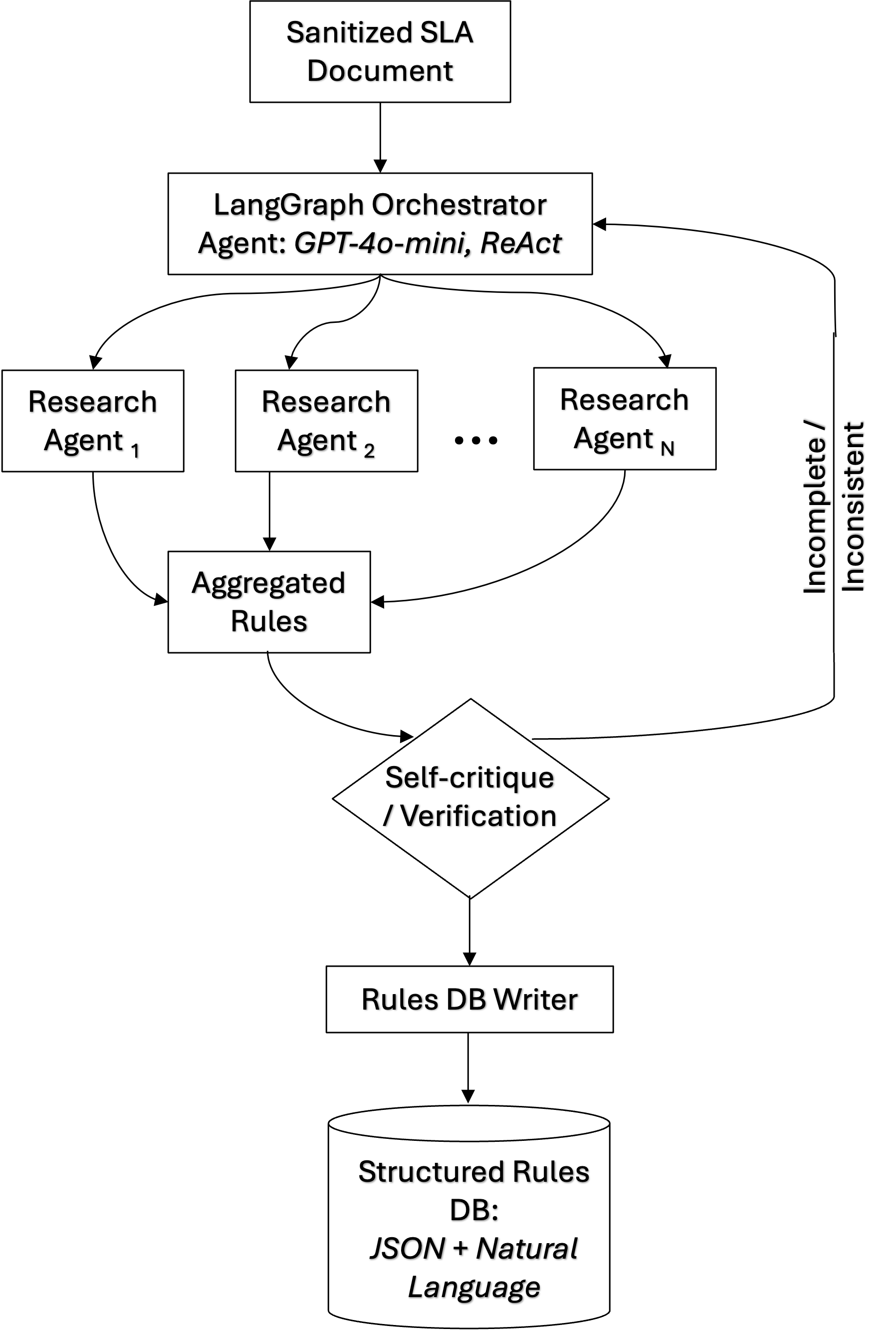}
    \caption{ReAct Agentic framework for SLA Rules Extraction into rules DB}
    \label{fig:system2}
\end{figure}

Shown in Fig.~\ref{fig:system2} is the framework for using the PII sanitized SLA document data into consistent, executable rule sets for each customer and contract into a rules DB. 
Conceptually, the ReAct framework system is structured around a LangGraph \emph{orchestrator agent} and a pool of $N$-specialized \emph{researcher agents}.
Each agent utilizes \texttt{GPT-4o-mini} model based on its role (see Fig.~\ref{fig:system2}).
The orchestrator is the single entry point receiving sanitized SLA documents from upstream, and maintains a shared state consisting of the current rules hypotheses associated with the received SLA, uncovered sections, and open questions.
It also has controls for when and how each researcher agent is invoked.

LangGraph is used to model this in the form of a graph with nodes for each agent, and edges to encode transitions with explicit control over the workflow, rather than a chain of prompts \cite{langgraph2024}.
Since SLA rule extraction can be posed as a stateful, multi-agent, iterative reasoning problem, LangGraph usage is justified.
The graph-based orchestration, loops, and state management in Fig.~\ref{fig:system2} allows us to encode the orchestrator-researcher-verification workflow explicitly, making the pipeline easier to control, extend, and debug over linear or prompt-chain-based alternatives.

To this end, we utilize a ReAct workflow where the orchestrator agent controls the flow, while triggering subordinate research agents with clear responsibilities to infer SLA rules of specific nature \cite{yao2022react}.
The orchestrator routes relevant excerpted text, and adds a context to each researcher, triggering the needed number of research agents in parallel \cite{shinn2023reflexion, schick2023toolformer}.
Each research agent then executes its core ReAct loop locally. 
It thinks through the assigned text, acts by emitting a structured candidate rule description in JSON + natural language, optionally requests follow-up snippets, and finally emits the structured JSON outputs.
The structured output is defined as the specific JSON schema with \texttt{\{metric, thresholds, SLA-tier, violation\_impact, comment-text\}}.
The candidate SLA rules are then aggregated into an "inferred rules set", capturing the beliefs of each research agent, but at this step they could be overlapping, or incomplete.

The orchestrator then triggers self-critique/ verification phase, central to our ReAct paradigm. 
It alternates between reasoning about the global rule set (e.g., "are all rules complete?", "are metrics present in all JSON schemas?", etc.) and re-invokes research agents again for incomplete or inconsistent items present in the inferred rules set. 
This forms a decision node in the graph.
The ReAct loop described above extracts interpretable and robust SLA rules complying with the specified structured schemas, that is more robust than single LLM calls, and requires multi-step reasoning.
As a result, it also consumes substantially more tokens than singular LLM calls.
However, this is justified as SLA documents that are parsed through the framework into our structured rules DB, do not need another parsing through the agentic workflow unless the SLA terms are modified again.
In real colocation operation settings, this occurs only at the end of the SLA life cycle (weeks to months).

At this point, we have obtained the structured rules DB (per customer, per SLA contract), which is an auditable single source of truth for the rest of the systems downstream.
An example result of one such structured rules instance is shown in Table.~\ref{tab:sla-rules-json}.
The remaining systems utilize past telemetry data and the rules DB to programmatically label sensor data, in order to train a multi-head attention model for predicting SLA violations.

\section{Multi-head attention models for colocation SLAs}\label{sec:attention}

Each customer's SLA is associated with specific racks at the colocation data center.
At this point, we start from two inputs for each customer: historical time-series of sensor data for the customer's GPU racks (e.g., power drawn, temperature, humidity), and contract-specific rules from the structured rules DB stored as JSON fields.
The historical telemetry is then normalized, resampled, and segmented into fixed‑length windows (overlapping 30‑minute look‑back sequences) so that each example captures enough temporal context for short‑horizon forecasting.

\begin{figure}[h!]
    \centering
    \includegraphics[width=1\columnwidth]{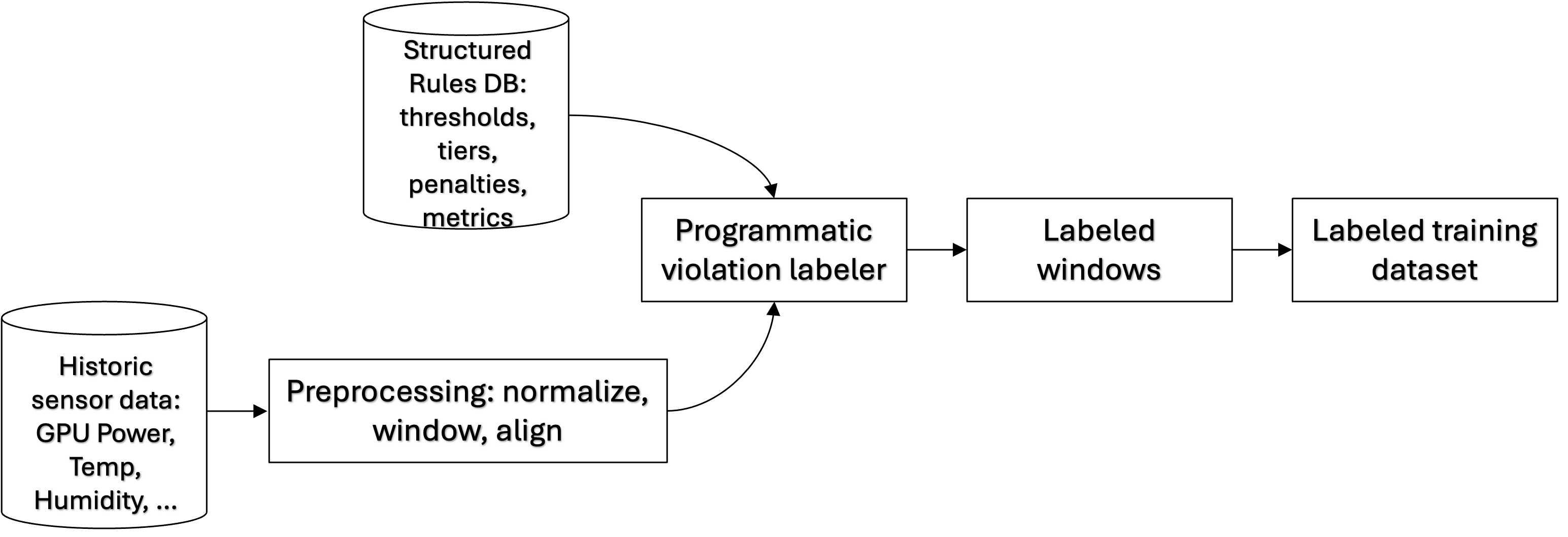}
        \vspace{-4mm}
\caption{Sensor Data \& Programmatic Labeling}
    \label{fig:system3}
\end{figure}

The programmatic labeling in Fig.~\ref{fig:system3} then applies the JSON rules deterministically to each window to derive labels per SLA rules DB entry.
For instance, for a power-drawn rule, we compute the aggregated power drawn at the rack level, compare it against the thresholds under rules DB, and assign one of the following labels \texttt{\{none, L1, L2\}}, based on which rules band the aggregate falls in.
Thus, instead of hand-labeling, every window yields a set of possible labels associated with the rules threshold, and aligns itself with one of the attention heads. 
Since all labels are generated directly from the rules used at inference time, labeling is auditable and easy to adapt upon modifications in the contract.
This can be done by simply updating the JSON schema upstream and re-running the labeler across a consistent training set.

Transformers are well suited for SLA violation prediction as they are often used in literature to model long-range temporal dependencies via self-attention, with no recurrence relations \cite{vaswani2017attention}.
Multi-headed attention also supports learning different views of similar temporal sequences, aligning with our design of dedicating each head to a separate SLA rule type.
More recent surveys in sequence learning have highlighted transformers as a key model family for capturing long-term temporal structures in multivariate settings \cite{kim2025comprehensive, zhang2024multi}.
Crucially, these models remain moderately interpretable when processing the diverse, dynamic variables found in data centers telemetry, especially when used with multi-headed attention.

\begin{figure}[h!]
    \centering
\includegraphics[width=1\columnwidth]{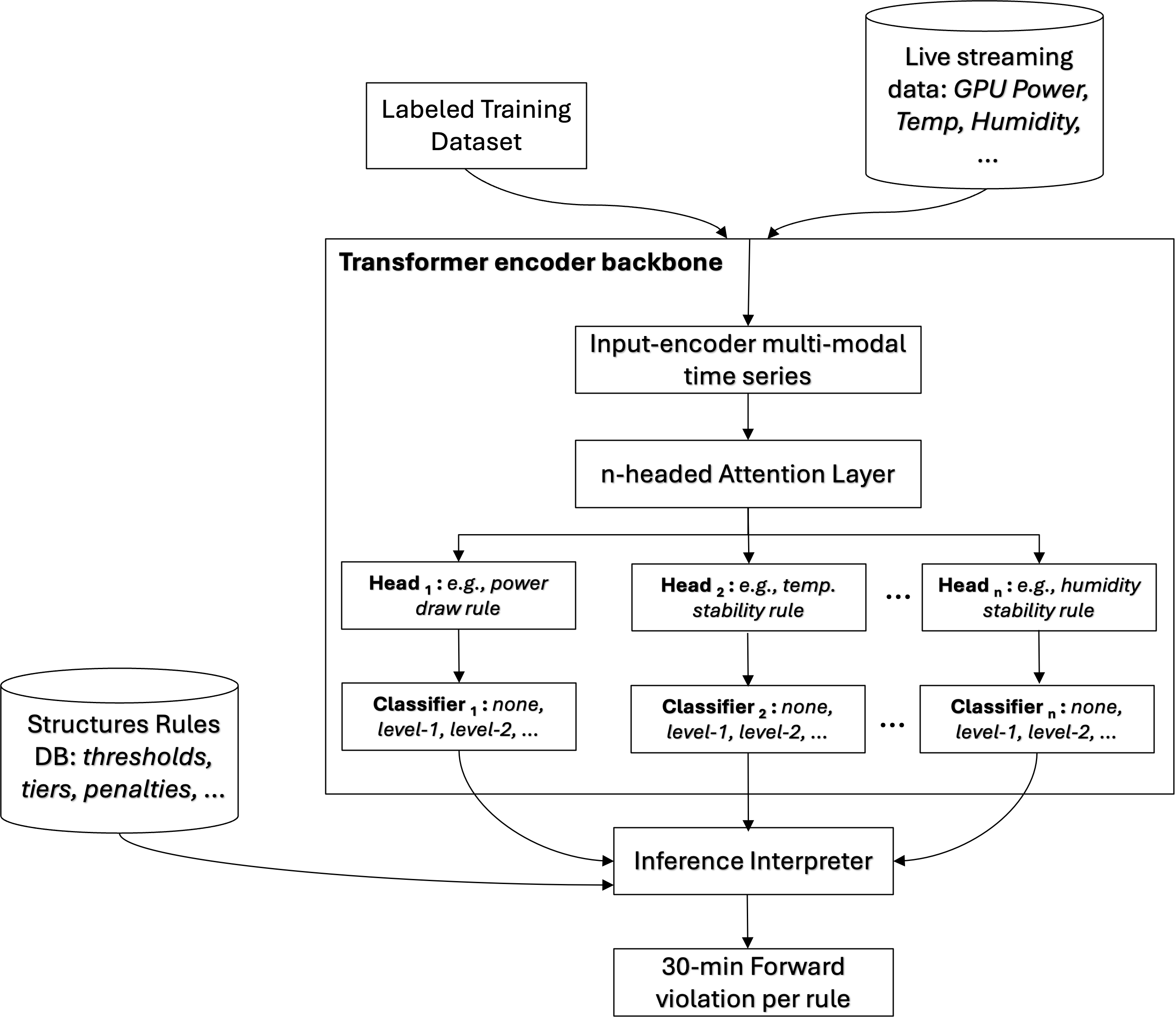}
    \caption{Per-Customer Transformer model with $n$-Attention Heads}
    \label{fig:system4}
\end{figure}

From an example JSON similar to Table.~\ref{tab:sla-rules-json}, we train our multi-headed transformer encoder on the labeled windows, treating each sequence of sensor readings as a multivariate time series classification problem.
Each time step in a window is embedded to include positional and temporal information.
For the example from Table~\ref{tab:sla-rules-json}, each embedded temporal window is passed through several layers of self-attention and feed-forward blocks, and then summarized into a latent sequence representation. 
On top of this shared backbone is an $n$-headed attention block.
In this example, it is a 3-headed attention block where Head-1 specializes in contracted power-drawn rule, Head-2 in the temperature stability rule, and Head-3 in humidity stability rule.
Each then feeds into its own classifier to produce a probability distribution over \{none, L1, L2\}. for the specific rule.
The overall schematic for the per-customer multi-head attention model is shown in Fig.~\ref{fig:system4}.

After training, each multi-head attention model is deployed as a streaming inference service to ingest continuous live telemetry data, and emits structured prediction events for every 30-minute rolling prediction window.
For each SLA rule (one per head), the service outputs predicted violation levels (None, L1, L2), associated probabilities, and a compact summary of the underlying sensor statistics in a JSON schema.
This becomes the canonical record for downstream reporting to relevant stakeholders.

\begin{figure}[h!]
    \centering
    \vspace{-2mm}
\includegraphics[width=1\columnwidth]{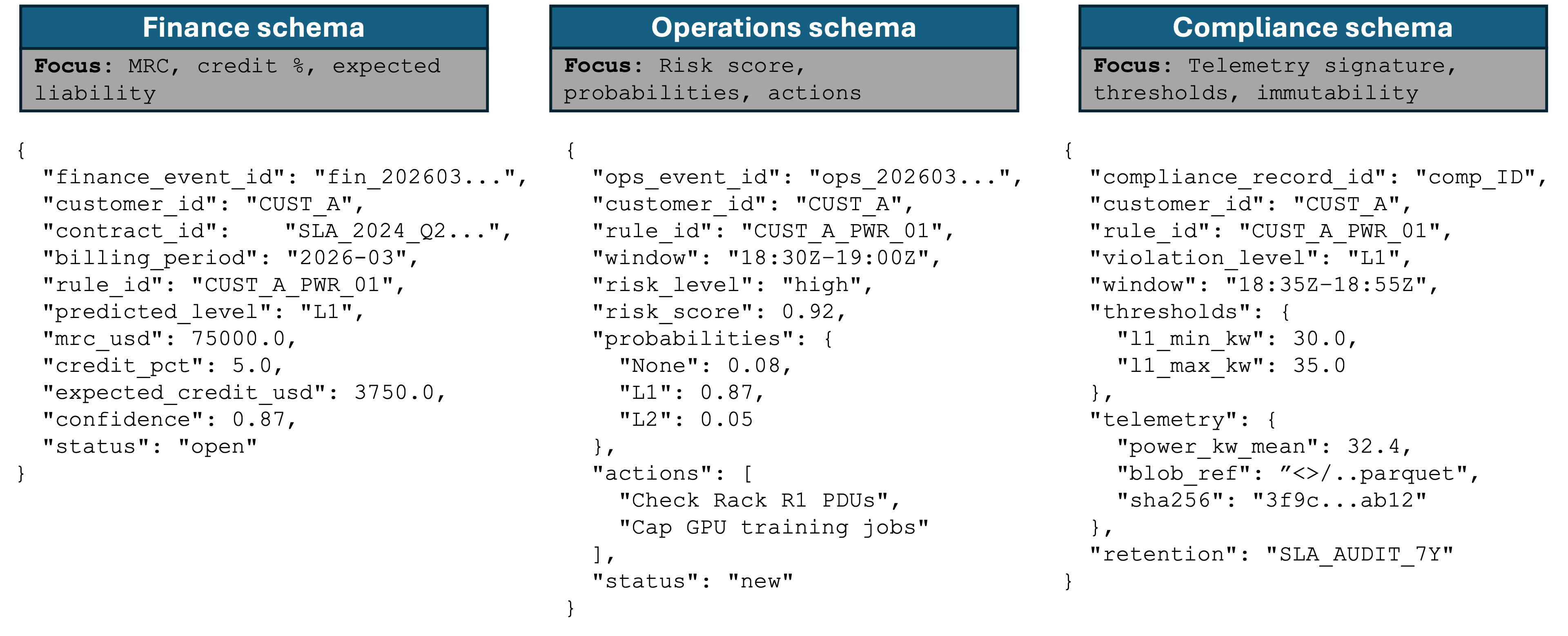}
    \vspace{-2mm}
    \caption{Post-Training Output Schemas for Stakeholders}
    \label{fig:out-schemas}
\end{figure}

For a real live system deployment, we define 3 different types of output schemas for each of the following stakeholder types: financial, operational, and compliance stakeholders.
One such output example of stakeholder specific schemas is shown in Fig.\ref{fig:out-schemas}.
This allows us to provide role-specific metrics and audit-ready evidence to stakeholders.
This structured, post-training pipeline turns model outputs directly into consumable artifacts for colocation business use case.

\begin{table*}[t!]
\centering
\footnotesize 
\caption{Example SLA rule encoding in the JSON Rules DB (SLA credit penalties as percentage of Monthly Recurring Charge).}
\label{tab:sla-rules-json}
\begin{tabular}{p{2cm}p{2.2cm}p{1.5cm}p{1.5cm}p{2cm}p{2.25cm}p{2.25cm}}
\hline
\textbf{Rule ID} & \textbf{Metric} [unit] & \textbf{Aggregation Window} & \textbf{Thresholds (None)} & \textbf{Thresholds (Level 1)} & \textbf{Thresholds (Level 2)} & \textbf{SLA Credit Penalty} \\
\hline
\noalign{\vspace{1ex}}
\texttt{CUST\_A\_PWR\_01}  & Power drawn [kW] & 5-min avg & $\leq 30$ & $30 < P \leq 35$ & $P > 35$ & None: 0\%; L1: 5\%; L2: 15\% \\
\texttt{CUST\_A\_TEMP\_01} & Temperature [$^\circ\mathrm{C}$] & 15-min avg & $18 \leq T \leq 27\,$ & $27 < T \leq 29$ or $16 \leq T < 18$ & $T < 16$ or $T > 29$ & None: 0\%; L1: 3\%; L2: 10\% \\
\texttt{CUST\_A\_HUM\_01}  & Humidity [\%RH]  & 15-min avg & $40 \leq H \leq 60\ $ & $35 \leq H < 40$ or $60 < H \leq 65$ & $H < 35$ or $H > 65$ & None: 0\%; L1: 2\%; L2: 8\% \\
\hline
\end{tabular}
\end{table*}

\begin{table*}[t!]
\centering
\footnotesize 
\caption{Mapping between multi-head attention outputs and SLA rules.}
\label{tab:head-rule-mapping}
\begin{tabular}{llllll}
\hline
\textbf{Head} & \textbf{Head Name}     & \textbf{SLA Rule ID} & \textbf{Metric}      & \textbf{Target Labels}        & \textbf{JSON Fields Used} \\
\hline
1 & Power head        & \texttt{CUST\_A\_PWR\_01}  & Power draw    & None / Level 1 / Level 2 & metric, window, thresholds, credits \\
2 & Temperature head  & \texttt{CUST\_A\_TEMP\_01} & Temperature   & None / Level 1 / Level 2 & metric, window, thresholds          \\
3 & Humidity head     & \texttt{CUST\_A\_HUM\_01}  & Humidity      & None / Level 1 / Level 2 & metric, window, thresholds          \\
\hline
\end{tabular}
\end{table*}

\section{System Demo}\label{sec:demo}
In this section we demonstrate our deployment build-out for the solution described above. 
We demonstrate the end-to-end pipeline on a semi-simulated multi-tenant colocation environment using customer-provided, proprietary power, temperature, and humidity telemetry from GPU racks, with SLA rules encoded per Table \ref{tab:sla-rules-json}. 
What follows is a fully functional system demonstration on this data. 
The system executes four sequential stages: SLA ingestion and PII scrubbing, ReAct-based rules extraction into the structured DB, programmatic data labeling, and transformer-based violation prediction with stakeholder output generation.

\subsection{Attention Model}
The per-customer multi-head transformer model totals 99,012 trainable parameters and is trained over 25 epochs using a 60-step look-back window targeting a 90-step prediction horizon. 
Training converges by the tenth epoch. Post-training evaluation exposes the complementary performance characteristics that motivate the decoupled attention head design. 

The multi-headed attention architecture successfully decoupled the heterogeneous facility metrics, allowing independent evaluation of power, temperature, and humidity dynamics. During the training phase, the model demonstrated robust predictive accuracy across all service-level agreement (SLA) tiers. 
By assigning independent attention heads to specific metric types, the system effectively mitigated cross-feature interference, maintaining high precision when predicting Level 1 (warning) and Level 2 (breach) events. 
The continuous ReAct agent loop ensured that as prediction confidence crossed predefined thresholds, the system could reliably trigger downstream alerts with minimal false positives, validating the framework's stability under simulated dynamic thermal and power load shifts.
Sample training data is summarized in the Fig.~\ref{fig:data} below. 

\begin{figure}[h!]
    \centering
\includegraphics[width=1\columnwidth]{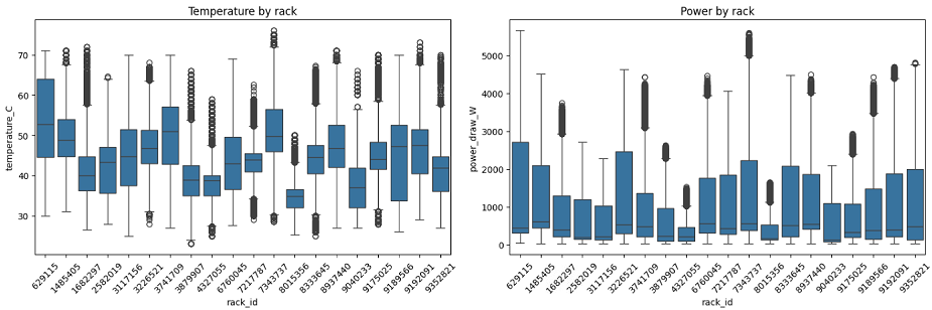}
\includegraphics[width=0.65\columnwidth]{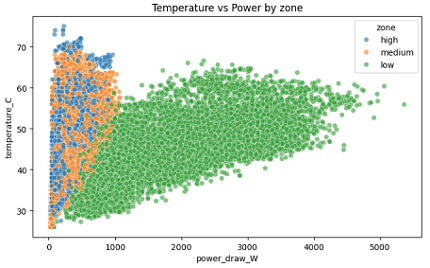}
\caption{Data Summaries for Sample Temperature and Power data used to train the transformer model (proprietary data PII removed); \textit{(above)} multi-tenant rack power \& ranges, \textit{(below)} typical rack temperatures measured at 3 points in colo racks.}
    \label{fig:data}
\end{figure}

\begin{figure}[h!]
    \centering
\includegraphics[width=1\columnwidth]{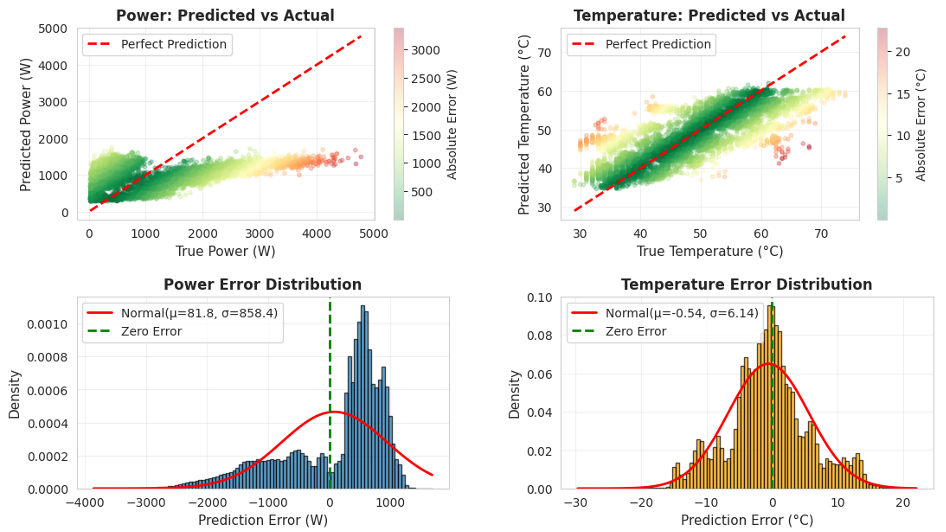}
    \caption{Model training results, and post training accuracy for power and temperature prediction}
    \label{fig:errors}
\end{figure}

The temperature head produces a near-Gaussian error distribution ($\mu$ = -0.54$^{\circ}$C, $\sigma$ = 6.14$^{\circ}$C) with a tight linear fit between predicted and actual values. 
The power head exhibits a skewed, bimodal error profile ($\mu$ = 81.8 W, $\sigma$ = 858.4 W) with systematic underprediction during high-load transients, a direct consequence of instantaneous load volatility (see Fig.~\ref{fig:errors}).
Critically, because each head learns an independent feature subspace, power transients do not degrade temperature prediction stability. This validates the architectural choice of strict 1-to-1 head-to-SLA-rule mapping.

\subsection{System Dashboard}

At inference time, the model ingests live telemetry over a rolling 30-minute window and emits a structured prediction event per SLA rule per head: a violation level (None, L1, or L2), associated probabilities, and a compact sensor statistics summary. 
The orchestration layer routes these events into three role-specific JSON schemas, shown in Fig.~\ref{fig:out-schemas}. 
The Finance schema maps the predicted violation to the contract's Monthly Recurring Charge, computing \texttt{credit\_pct} (e.g., 5.0\% for a Level 1 power breach) and \texttt{expected\_credit\_usd} for real-time revenue-at-risk tracking. 
The Operations schema translates model confidence intervals into a categorical \texttt{risk\_level} and surfaces contextual interventions such as ``Check Rack R1 PDUs" or ``Cap GPU training jobs." 

\begin{figure}[h!]
    \centering
    \includegraphics[width=1\columnwidth]{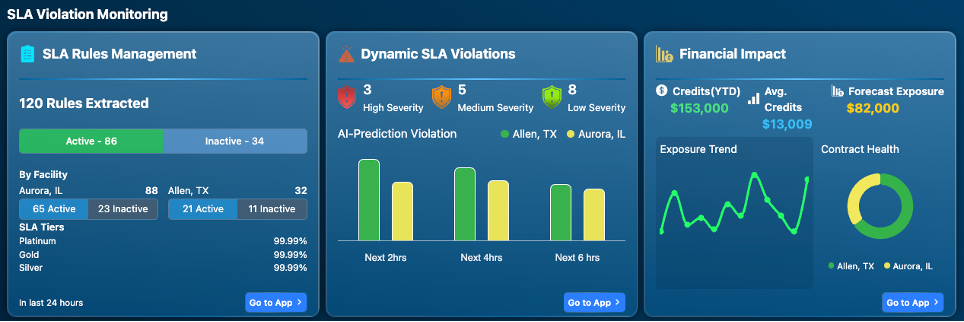}
\includegraphics[width=1\columnwidth]{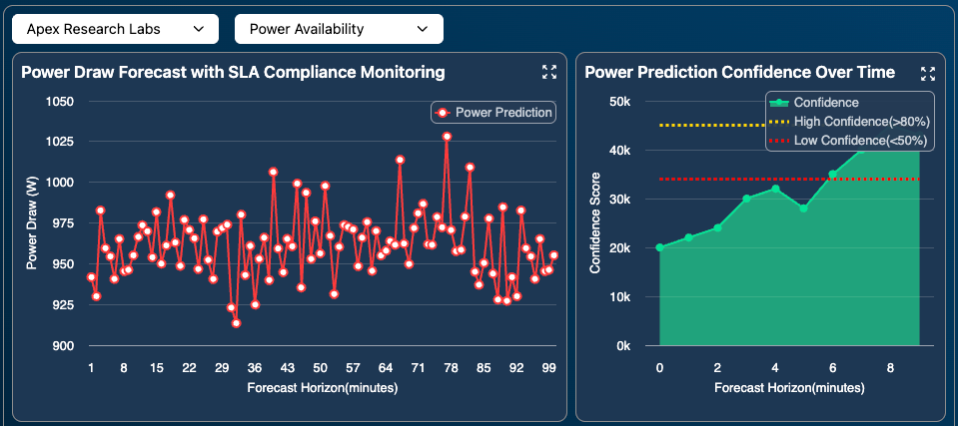}
    \caption{Deployment Dashboard for monitoring SLA Violations: \textit{(above)} Rules Management, Inference functionality using ReAct framework, \textit{(below)} Dynamic SLA Violation prediction for a selected power metric.}
    \label{fig:dashboard}
\end{figure}

These structured outputs feed a unified SLA Dashboard, shown in Fig.~\ref{fig:dashboard}, which aggregates operational risk scores and projected financial liabilities across all tenants in a single view. 
Facility operators can monitor simultaneous, independent SLA compliance states in real-time without interpreting raw attention weights or log files, directly operationalizing the predictive model's output.

\section{Conclusion}\label{sec:conclusion}
In this work we presented our industry-validated, end-to-end framework to automate service level agreement compliance monitoring in multi-tenant colocation data center facilities.
We combined a ReAct-based agentic rules extraction pipeline with a per-customer multi-head transformer model to address the industry-validated gap: existing colocation reliance on manual SLA management and post-hoc breach analysis.
We shift this gap towards a proactive model by encoding SLA contracts as machine-readable JSON rules, programmatically labeled historical telemetry, and trained light-weight transformer models (\~90k parameters) to predict violations up to 90 time steps ahead.
Our post-training  system generated finance, operations, and compliance output schemas to convert model predictions into operational risk estimates.

Our future steps include maintaining auditable SLA traces, evidence backed SLA reports, and standardized structured rules databases by involving AI agents in a stricter output schema-based framework.
Our models are to be fine-tuned for real-world multi-facility colocation data center operations and deployed.

\bibliographystyle{IEEEtran}
\bibliography{bibliography}

\end{document}